%% file: neurips_2024.tex
\documentclass{article}

\PassOptionsToPackage{numbers, compress}{natbib}
\usepackage[preprint]{neurips_2024}

\usepackage[utf8]{inputenc} 
\usepackage[T1]{fontenc}    
\usepackage{hyperref}       
\usepackage{url}            
\usepackage{booktabs}       
\usepackage{amsfonts}       
\usepackage{nicefrac}       
\usepackage{microtype}      
\usepackage{xcolor}         

\usepackage[pdftex]{graphicx}
\usepackage{caption}
\usepackage{subcaption}
\graphicspath{ {./images/} }
\usepackage{amsmath}
\usepackage{colortbl}

\usepackage{comment}

\newcommand{\vect}[1]{\boldsymbol{#1}}

\definecolor{best}{rgb}{0.96, 0.57, 0.58}
\definecolor{second}{rgb}{0.98, 0.78, 0.57}
\definecolor{third}{rgb}{1.0, 1.0, 0.56}

\title{Normal-GS: 3D Gaussian Splatting with Normal-Involved Rendering}

\author{%
  Meng Wei$^{1}$ \quad Qianyi Wu$^{1}$ \quad Jianmin Zheng$^2$ \quad Hamid Rezatofighi$^{1}$ \quad Jianfei Cai$^1$ \\
  $^1$Monash Univeristy \quad $^2$Nanyang Technological University \\
  \texttt{\{meng.wei,qianyi.wu,hamid.rezatofighi,jianfei.cai\}@monash.edu} \\
  \texttt{\{ASJMZheng\}@ntu.edu.sg}
}

\begin{document}
  \maketitle

\input{Section/0_abs}
\input{Section/1_intro}
\input{Section/2_related}
\input{Section/3_method}
\input{Section/4_experiment}
\input{Section/5_conclusion}

  {
    \small

    \bibliography{cite}
  \end{document}

%% file: Section/0_abs.tex
    \begin{abstract}
    Rendering and reconstruction are long-standing topics in computer vision and graphics. Achieving both high rendering quality and accurate geometry is a challenge. Recent advancements in 3D Gaussian Splatting (3DGS) have enabled high-fidelity novel view synthesis at real-time speeds. However, the noisy and discrete nature of 3D Gaussian primitives hinders accurate surface estimation. Previous attempts to regularize 3D Gaussian normals often degrade rendering quality due to the fundamental disconnect between normal vectors and the rendering pipeline in 3DGS-based methods. Therefore, 
    we introduce \textbf{Normal-GS}, a novel approach that integrates normal vectors into the 3DGS rendering pipeline. The core idea is to model the interaction between normals and incident lighting using the physically-based rendering equation.
    Our approach re-parameterizes surface colors as the product of normals and a designed Integrated Directional Illumination Vector (IDIV).
    To save memory usage and simplify the optimization, we employ an anchor-based 3DGS to implicitly encode locally-shared IDIVs.
    Additionally, Normal-GS leverages optimized normals and Integrated Directional Encoding (IDE) to accurately model specular effects, enhancing both rendering quality and surface normal precision.
    Extensive experiments demonstrate that Normal-GS achieves near state-of-the-art visual quality while obtaining accurate surface normals and preserving real-time rendering performance.
  \end{abstract}

%% file: Section/1_intro.tex
  \section{Introduction}
  \label{introduction}

Radiance Fields have emerged as a prominent representation in 3D vision, largely propelled by the pioneering advancements of Neural Radiance Fields (NeRF)~\cite{mildenhall2021nerf}. Despite their success, NeRFs are hindered by prolonged rendering times and cumbersome training processes. Recently, 3D Gaussian Splatting (3DGS) has been introduced to represent Radiance Fields using millions of 3D Gaussians,
each endowed with additional attributes such as opacity and color~\cite{3DGS_2023}. 3DGS significantly enhances rendering efficiency through CUDA-accelerated rasterization~\cite{EWAvolumeSplatting}, achieving comparable or superior fidelity to NeRF-based models. The
advantages of 3DGS have garnered significant attention~\cite{wu2024recent}, prompting numerous studies aiming at enhancing its rendering capabilities~\cite{yang2024specgaussian,scaffoldgs,jiang2023gaussianshader} and improving the quality of its underlying geometry~\cite{guedon2023sugar, cheng2024gaussianpro,Dai2024GaussianSurfels}.

However, a critical issue continues to plague the development of 3DGS: \emph{the quality of appearance and geometry seemingly oscillates like a seesaw.} Efforts to refine the geometry often involve the incorporation of regularization techniques~\cite{guedon2023sugar, chen2023neusg} or additional geometry supervision~\cite{li2024dngaussian, turkulainen2024dnsplatter}, which may compromise rendering fidelity. Conversely, advancements in appearance modeling, such as the integration of implicit networks~\cite{scaffoldgs} and sophisticated shading attributes~\cite{yang2024specgaussian,jiang2023gaussianshader}, struggle to concurrently elevate both appearance and geometry quality. This raises a pertinent question: \emph{Is it feasible to achieve a high-quality appearance while also capturing precise underlying geometry information in 3DGS?}

Revisiting the classical rendering equation~\cite{kajiya1986rendering}, we identify a fundamental issue in the appearance-geometry conflict: the disconnection between surface normals and the rendering process in current 3DGS methods. 
Specifically, the existing methods render pixel colors by alpha-blending colored 3D Gaussians. The color of each Gaussian is queried by the function value among Spherical Harmonics~\cite{hobson1931theory} along the ray direction,
which disregards the contributions of surface normals. This oversight hampers the ability to achieve a balanced integration of appearance and geometry. 
Our framework draws inspiration from the conventional rendering equation, which incorporates surface normals into appearance calculations.
Although prior attempts have explored this integration in contexts such as inverse rendering~\cite{liang2023gs,R3DG2023} and specular appearance modeling~\cite{jiang2023gaussianshader,yang2024specgaussian},
they leverage simplified model~\cite{burley2012physically} or approximation approaches like split-sum~\cite{karis2013real} or global environmental map~\cite{greene1986environment,ramamoorthi2001efficient} to decompose appearance into several individual components with meticulous regularization terms, which either compromising in rendering quality or geometric accuracy.
This challenge highlights the complexity of balancing various components for physically-based rendering, prompting our design of a straightforward method to integrate normal information explicitly and simply into Gaussian appearance modeling.

\begin{figure}
    \centering
    \includegraphics[width=\linewidth]{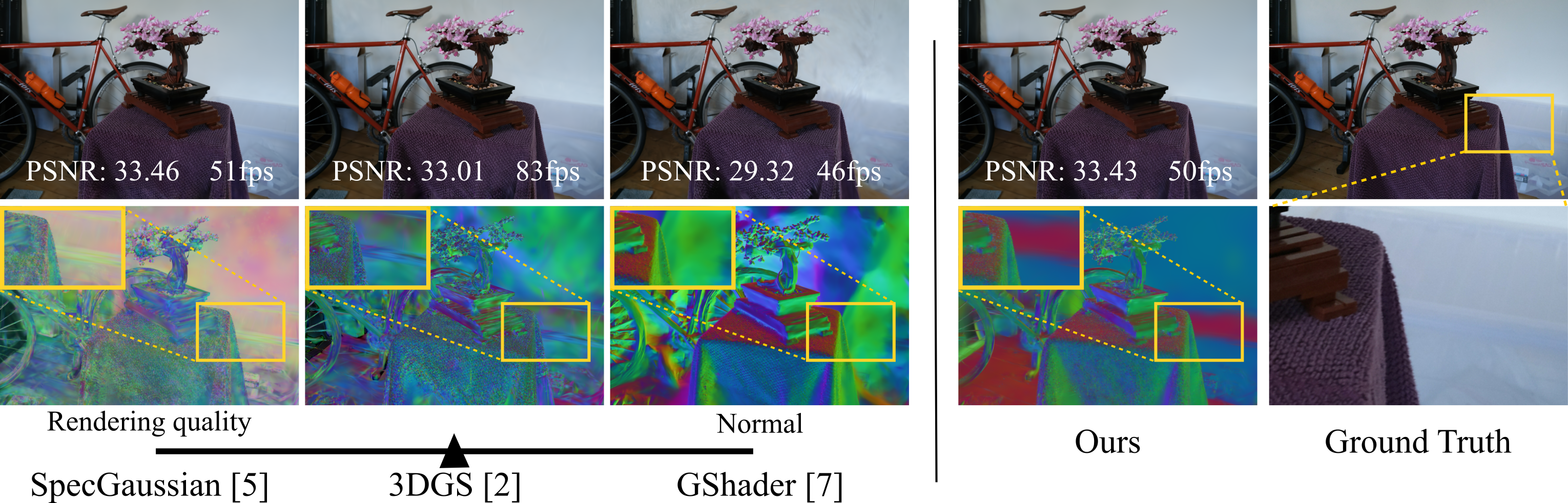}
    \caption{The "seesaw" characteristic between the rendering quality and the normal accuracy of 3DGS-based methods. Our \textbf{Normal-GS} is able to 
    efficiently achieve accurate normal estimation while preserving competitive rendering quality. Our method successfully captures the normals of the cover of the semi-transparent box behind the table.}
    \label{fig:first}
    \vspace{-3mm}
  \end{figure}

In this work, we propose a normal-involved shading for 3DGS, called \textbf{Normal-GS}, for high-quality rendering and accurate normal estimation, while maintaining real-time performance. Our method directly accounts for normal contributions in the 3DGS rendering pipeline, enabling gradient signals to backpropagate properly to better align 3D Gaussians. 
Specifically, Normal-GS considers the normal and incident lighting interactions at the surface from the physically-based shading perspective and explicitly models normals' contributions for diffuse and specular components. By considering 
the low-frequency nature of the incident light field, we further implicitly model it 
with MLPs. 
With our strategy, compared with the state-of-the-art (SOTA) 3DGS methods, we achieve competitive rendering quality, more accurate normal information, and real-time rendering, see one example in~Fig.~\ref{fig:first}.
Our contributions are summarized below.
\begin{itemize}
    \item We propose a new formulation to involve normal information in appearance modeling for 3DGS, which strives for a good balance between appearance and geometry modeling.
    \item We leverage an effective structure to regularize our framework using anchor-based MLPs, without introducing many extra regularization terms and achieving better quality.
    \item Extensive experiments demonstrate our framework for obtaining superior view synthesis and normal estimation. Our design could serve as a plug-in component for 3DGS approaches.
\end{itemize}

%% file: Section/2_related.tex
  \section{Related Works}
  \label{related}

  \paragraph{Radiance Fields: From Neural Radiance Fields to 3D Gaussian Splatting.}
Neural Radiance Field (NeRF) has significantly influenced the field of 3D vision with its impressive novel view synthesis capabilities \cite{mildenhall2021nerf}. NeRF catalyzed advancements in neural rendering, inspiring a multitude of subsequent research focused on enhancing rendering quality under complex lighting and material conditions \cite{verbin2022ref,yao2022neilf,zhang2023neilf++}, varying camera distributions \cite{barron2021mip,barron2022mipnerf360,hu2023tri}, and scalability for large scenes \cite{turki2022mega,wang2023f2,barron2023zip}. However, the implicit representation used in NeRF, which relies on a Multi-layer Perceptron (MLP) to compute density and radiance for any given 3D position and ray direction, necessitates extensive computations. This inefficiency has spurred efforts to enhance the practicality of NeRF, with notable strides made in accelerating its processing efficiency \cite{3DGS_2023,yu2021plenoctrees,muller2022instant}.

Among these advancements, 3D Gaussian Splatting (3DGS) has emerged as a promising solution \cite{3DGS_2023}. By explicitly modeling 3D scenes using sets of 3D Gaussians and employing tile-based GPU rasterization, 3DGS achieves real-time rendering with competitive quality. Its success has spurred applications in areas like 3D generation \cite{tang2023dreamgaussian,yi2023gaussiandreamer}, physical simulation \cite{xie2023physgaussian}, and sparse view reconstruction \cite{charatan2023pixelsplat,chen2024mvsplat}, despite its higher memory requirements for scene representation. Innovations such as feature anchoring in Scaffold-GS \cite{scaffoldgs}, value pruning and quantization techniques~\cite{Joo,zhiwen,Simon} and mining spatial relationship~\cite{Wieland,hac2024,yang2024spectrally} are among the efforts to reduce the memory footprint of 3DGS, thereby enhancing storage efficiency and rendering fidelity.
\paragraph{Geometry and Appearance in 3DGS: From Surface Reconstruction to Inverse Rendering.}
Geometry and appearance are central to 3D reconstruction, and the discrete and explicit nature of 3DGS can result in noisy underlying geometry~\cite{guedon2023sugar,Dai2024GaussianSurfels,chen2023neusg,turkulainen2024dnsplatter,huang20242d}. SuGaR \cite{guedon2023sugar} was an initial attempt to refine mesh extraction from 3DGS, applying regularization to align the Gaussian Splatting with the actual surface contours. To better define geometry, capturing normal information is essential, leading to innovations like transforming 3DGS into 2D Gaussian Splatting (2DGS) and Gaussian Surfels \cite{Dai2024GaussianSurfels,huang20242d}. These methods apply regularization to depth and normal rendering, ensuring that the Gaussians are appropriately distributed across surfaces. While these approaches improve surface reconstruction, they often sacrifice rendering fidelity in novel view synthesis due to the lack of a clear relationship between geometry and appearance \cite{jiang2023gaussianshader, Dai2024GaussianSurfels,huang20242d}.

In appearance modeling, the focus has shifted to inverse rendering for 3DGS, which aims to separate scene elements into materials, lighting, and geometries. This inherently unconstrained problem challenges traditional rendering equations. Techniques such as Monte Carlo integration, point-based ray tracing \cite{R3DG2023, shi2023gir}, and baking \cite{liang2023gs} are employed to handle the complex integrals, often relying on simplified models like the Disney Bidirectional Reflectance Distribution Function (BRDF) \cite{burley2012physically} and approximation methods such as split-sum \cite{liang2023gs,R3DG2023,shi2023gir,ye20243d} or environmental mapping \cite{jiang2023gaussianshader,ye20243d,wu2024deferredgs}. Despite their efforts, these simplifications generally result in lower rendering quality, as reflected in reduced Peak Signal-to-Noise Ratios (PSNR) \cite{liang2023gs,R3DG2023,shi2023gir}, failing to match the original 3DGS.

It is worth noting that some concurrent works~\cite{wu2024deferredgs,yu2024gsdf} have a similar motivation as ours.~\cite{yu2024gsdf} simultaneously enhance geometry accuracy and rendering quality by incorporating a laborious dual-branch framework, which uses 3DGS for appearance rendering and an implicit neural surface for geometry production. While this approach integrates the strengths of each system, it is hampered by slower training speeds relative to standalone 3DGS, 
due to its additional networks for surface representation. 
~\cite{wu2024deferredgs} also distills geometry information from an extra neural implicit surface network for deferred rendering. Our solution proposes a new perspective from a rendering standpoint to consider geometry and appearance simultaneously without an extra implicit network for the surface.

%% file: Section/3_method.tex
\vspace{-0.2em}
\section{Method}
\vspace{-0.2em}
Our goal is to simultaneously enhance both image quality and normal estimation capability of 3DGS, while maintaining real-time rendering. To achieve this, we first analyze the existing rendering pipeline in 3DGS to identify its limitations in not involving surface normals in color modeling (Sec.~\ref{method:preliminary}).
We then design a normal-involved rendering strategy that aligns with the physically based rendering equation. Our method effectively models interactions between normals and incident lighting, parameterized by the proposed Integrated Directional Illumination Vectors (IDIV) (Sec.~\ref{method:oi}). We employ an anchor-based GS to implicitly encode locally shared IDIVs to save memory and aid optimization (Sec.~\ref{method:local}).
The training details of our framework are provided in Sec.~\ref{method:train_details}.

\subsection{Preliminary} \label{method:preliminary}
\paragraph{3D Gaussian Splatting \cite{3DGS_2023}.} 3DGS models scenes using a set of discrete 3D Gaussians, each defined by its spatial mean $\vect{\mu}$ and covariance matrix $\Sigma$:
\begin{equation}\label{eq:3Dgaussian}
  G(\vect{p}) = \exp(-\frac{1}{2} (\vect{p}-\vect{\mu})^T\Sigma^{-1}(\vect{p} -\vect{\mu})).
\end{equation}
The covariance matrix $\Sigma$ is parameterized by a scaling matrix $S$ and a rotation matrix $R$, such that $\Sigma = RSS^TR^T$, ensuring it remains positive semi-definite during optimization.

Each 3D Gaussian is associated with a color $c$ and an opacity $\alpha$. During rendering, these Gaussians are projected (rasterized) onto the image plane, forming 2D Gaussian splats $G'(x)$, as described in \cite{EWAvolumeSplatting}. The 2D Gaussian splats are sorted from front to back tile-wisely, and $\alpha$-blending \cite{inria2021point} is performed for each pixel $x$ to render its color as follows:
\begin{equation} \label{eq:3DGS}
  C(x) = \sum_{i\in N} c_i \sigma_i \prod_{j=1}^{i-1}(1-\sigma_j), \quad \sigma_i=\alpha_i G_i'(x)
\end{equation}
where $N$ specifies the number of 2D Gaussian splats covering the current pixel. Heuristic densification and pruning strategies \cite{3DGS_2023} are employed to address potential under- and over-reconstruction and ensure multi-view consistency in rendered images.

The color of each Gaussian, $c$, is represented by Spherical Harmonics (SH) as $k_l^m Y_l^m(\vect{\omega}_\textrm{view})$ to provide view-dependent effects, where $(l, m)$ is the degree and order of the SH basis $Y_l^m$, $k_l^m$ is the corresponding SH coefficient, and $\vect{\omega}_\textrm{view}$ specifies the viewing direction. 3DGS usually uses a maximum degree $l$ of 3, formulating $c(\vect{\omega_{\textrm{view}}}) = \sum_{l=0}^{3} \sum_{m=-l}^{l} k_l^m Y_l^m(\vect{\omega_{\textrm{view}}})$.

In Eq.~\eqref{eq:3DGS}, standard 3DGS treats colors as intrinsic attributes, 
independent of surface normals. This independence prevents surface normals from receiving gradient signals during the backpropagation pass when optimizing surface colors. Moreover, this separation significantly undermines our goal of simultaneously enhancing image quality and normal estimation capability, since improvements in normals cannot contribute to the rendering quality in the forward shading pass of 3DGS. 

\subsection{Normal-Involved Shading Strategy}
\label{method:oi}

\paragraph{Physically Based Rendering.}
To integrate surface normals into both the backward and forward rendering passes of 3DGS effectively, we propose a normal-involved rendering strategy. This strategy adopts principles from physically based surface rendering \cite{kajiya1986rendering} in computer graphics, which models the out radiance $L_\text{out}$ of a surface point as a function of the incident lighting $L_\textrm{in}$ and normals. Specifically, for each 3D point, its out-radiance in the outward direction $\vect{\omega_o} = -\vect{\omega_{\textrm{view}}}$ is defined as:
\begin{equation} \label{eq:kajiya}
  c(\vect{\omega_{\textrm{view}}}) = L_\text{out} (\vect{\omega_o}) = L_\textrm{E}(\vect{\omega_o}) + \int_{\Omega^+}  L_\textrm{in}(\vect{\omega_i}) (\vect{\omega_i}\cdot \vect{n}) f_r(\vect{\omega_i}, \vect{\omega_o}) \,d\vect{\omega_i}
\end{equation}
where $L_\textrm{E}$ is the emitted radiance along the outward direction, the integral is defined over the upper hemisphere $\Omega^+$ of the surface; $\vect{\omega_i}$ is the incident light direction; BRDF $f_r(\cdot)$ describes how light is reflected from $\vect{\omega_i}$ to $\vect{\omega_o}$; the cosine term $(\vect{\omega_i}\cdot \vect{n})$ accounts for the light geometric attenuation.

\paragraph{Integrated directional illumination vectors for Lambertian objects.}
We first consider a simple ideal case where the scene contains only Lambertian objects. For more complex materials, we handle them in Sec.~\ref{method:model}. Given that the diffuse reflection of Lambertian objects is view-independent, the complicated BRDF function $f_r(\cdot)$ can be reduced to a spatially varying albedo term $k_\textrm{D}$. Omitting the emitted radiance $L_\textrm{E}$, the rendering equation accounting for the diffuse reflectance $L_\textrm{D}$ becomes:
\begin{equation}
\label{eq:diffuse}
  L_\textrm{D} = \int_{\Omega^+} L_\textrm{in}(\vect{\omega_i})\cdot k_\textrm{D} \cdot (\vect{\omega_i}\cdot \vect{n}) \,d {\vect{\omega_i}}
  = k_\textrm{D} \int_{\Omega^+} L_\textrm{in}(\vect{\omega_i})  (\vect{\omega_i}\cdot \vect{n}) \,d {\vect{\omega_i}}.
\end{equation}

We want to find a way to explicitly re-parameterize $L_\textrm{D}$ as a function of normal $\vect{n}$, meanwhile avoiding complicated integrals to save time. Specifically, we want to represent $L_\textrm{D}$ as a dot product between $\vect{n}$ and some illumination vector $\vect{l}$. 
%
Inspired by the traditional shape-from-shading strategy \cite{xu2017shading}, we extract the normal out of Eq.~\eqref{eq:diffuse}:
\begin{equation}
  L_\textrm{D} = k_\textrm{D}\cdot \vect{n} \cdot \left[\int_{\Omega^+} L_\textrm{in}(\vect{\omega_i}) \vect{\omega_i} \,d {\vect{\omega_i}} \right].
\end{equation}
By defining a new term called ``\textit{integrated directional illumination vector} (IDIV)'' as
\begin{equation}
  \vect{l} = \int_{\Omega^+} L_\textrm{in}(\vect{\omega_i}) \vect{\omega_i} \,d {\vect{\omega_i}},
\end{equation}
we model the color for Lambertian objects as the albedo times a dot product between IDIV and the surface normal:
$c = L_\textrm{D} = k_\textrm{D} \cdot \vect{n} \cdot \vect{l}$. Compared with the original 3DGS and the method \cite{jiang2023gaussianshader} that directly use a constant diffuse color, our re-parameterization successfully accounts for the normal's contributions to the diffuse component of surface colors.

\paragraph{Analysis.}
Our re-parameterization of the diffuse color enables the normal vector to be computationally involved in the rendering pass. As such, during backpropagation, by the chain rule, the surface color could pass gradient signals to the normal vectors as follows
\begin{equation}\label{eq:bp_normal}
  \frac{d\mathcal{L}}{d\vect{n}} = \frac{d\mathcal{L}}{dc} \cdot  \frac{dc}{d\vect{n}} = \frac{d\mathcal{L}}{dc} \cdot (k_\textrm{D} \cdot \vect{l}).
\end{equation}

Other image-based methods \cite{jiang2023gaussianshader,liang2023gs, ye20243d} usually optimize a single environment map $E$ and query this map using normals to determine diffuse colors. Consequently, their gradients on normals function by rotating the 3D Gaussians until the queried environment map color matches the pixel color, making the optimization process heavily dependent on the quality of $E$. These methods assume the existence of a global environment map, which may not be valid for general real scenes. In contrast, our approach employs IDIVs to capture local incident lighting, eliminating the reliance on such a strong assumption and providing a more flexible and accurate framework for optimizing surface normals.

  \begin{figure}[t]
    \centering
    \includegraphics[width=\linewidth]{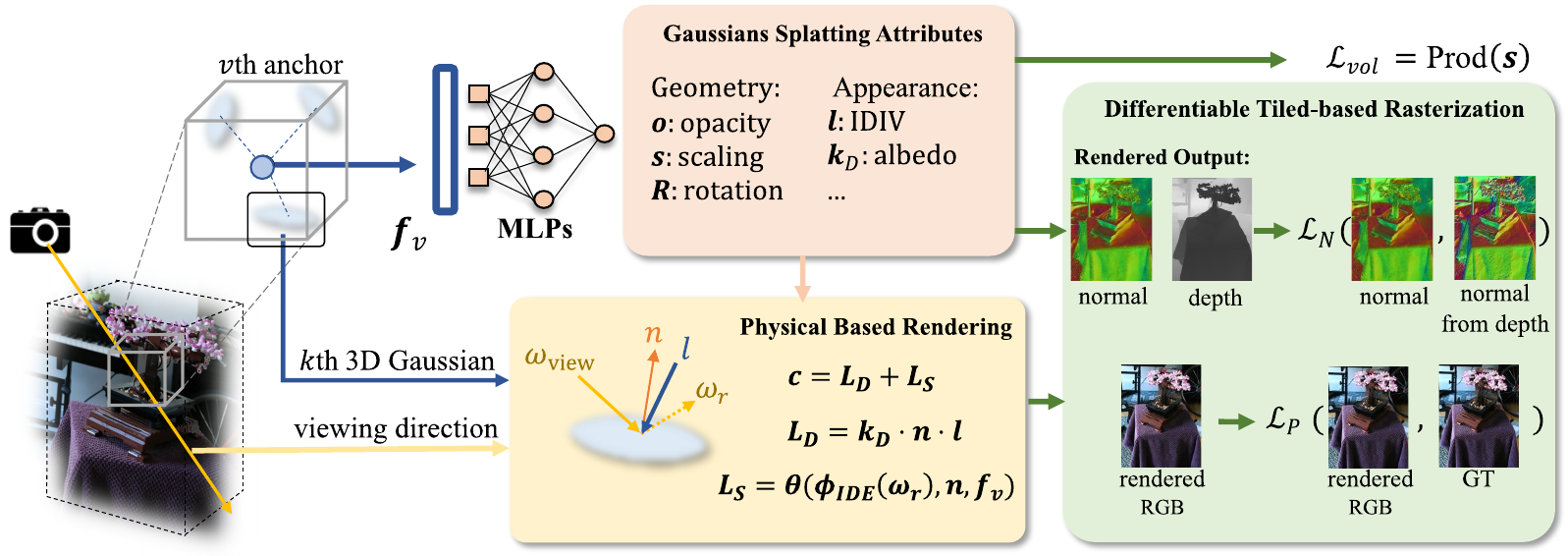}
    \caption{Our normal-involved GS, Normal-GS, reparameterizes the original colors into the diffuse and specular components, $c=L_\textrm{D} + L_S$ (bottom). It models the diffuse component as the dot product between the normal vector $\vect{n}$ and the Integrated Directional Illumination Vector (IDIV) $\vect{l}$, and utilizes the Integrated Directional Encoding (IDE) \cite{verbin2022ref} to capture view-dependent specular effects. Inherent parameters are encoded implicitly by a locally shared anchor Gaussian (left) and decoded using MLPs (top). Our method accounts for the contributions of normals to colors, effectively enhancing geometry accuracy and rendering quality.}  \label{fig:method}
  \end{figure}

  \subsection{Modeling Locally Shared Integrated Directional Illumination Vectors}
  \label{method:local}
  Replacing 3D Gaussian colors $c$ with $k_\textrm{D} \cdot \vect{n} \cdot \vect{l} $ introduces additional free parameters.
  Therefore, it is necessary to regularize the solution space of the introduced variables $k_\textrm{D}$ and $\vect{l}$ to stabilize the training process and avoid overfitting.

  \paragraph{Regularizing the solution space.} \citet{xu2017shading} address solution space constraints from an optimization perspective. They have introduced several regularizers, such as total variation and Laplacian terms, to enhance the performance. However, their mesh-based method relies on a connected topology, which is incompatible with the discrete and sparse nature of 3D Gaussians. Although k-NN can be used to find connected primitives, it severely hampers training speed. Moreover, since 3D Gaussians can be semi-transparent, directly applying their methods to our case is unsuitable.

  Instead, we approach the problem from a neural rendering perspective. As observed in \cite{scaffoldgs,hac2024,ye2023gaussian,ververas2024sags}, 3D Gaussians with similar textures usually cluster together. A similar situation applies to incident lighting, which tends to be locally shared and exhibits low-frequency variations.
  To avoid the redundancy of storing Integrated Directional Illumination Vectors (IDIVs) for each Gaussian and capture local coherence, we propose to encode IDIVs with locally shared features and decode them using MLPs. An additional advantage of using MLPs to approximate IDIVs is their inherent smoothness in function approximations, aligning with our objectives. 

  In particular, we employ an anchor-based Gaussian Splatting (GS) method, Scaffold-GS \cite{scaffoldgs}, to implicitly represent IDIVs using locally shared features $\vect{f}_v$ stored at anchors $v$ and decode them using a global MLP, $\theta_{l}(\vect{f}_v) = \{\vect{l}_v^k\}_{k=1}^{K}$, where $K$ IDIVs share a local feature $\vect{f}_v$. Therefore, we only need to learn a compact set of per-anchor local features and use the MLP to generate per-Gaussian IDIVs, significantly reducing the dimensionality of the problem. Further details on our model configurations can be found in the following Sec.~\ref{method:model} and Appendix.

  \subsection{Training Details}\label{method:train_details}
  Our method adopts the established 3DGS pipeline \cite{3DGS_2023}, which first performs Structure-from-Motion \cite{schoenberger2016sfm} on input images to generate sparse points. These points serve as initial 3D Gaussians and are optimized to model the scene.

  \paragraph{Defining surface normals on 3D Gaussians.}
  During optimization, 3D Gaussians often exhibit flatness around the surface areas, as observed in \cite{ yang2024specgaussian,jiang2023gaussianshader,shi2023gir}. For a near-flat Gaussian ellipsoid, its shortest axis thus functions as the normal vector.
  However, the shortest axis of 3D Gaussians does not physically represent the actual surface normal. In contrast, implicit radiance fields leverage the gradient of densities to naturally approximate surfaces, and the Signed Distance Function (SDF) has its gradients as surface normals by definition. To make the shortest axis of 3D Gaussians align with the real surface normal, we employ a depth regularized loss term on normals, similar to \cite{jiang2023gaussianshader,Dai2024GaussianSurfels,huang20242d}.

  We first render the depth and normal images, $\{\mathcal{D, N}\}$, using the 3DGS tile-based rasterizer, which is achieved by replacing the color term in Eq.~\ref{eq:3DGS} with depths and normals of 3D Gaussians, respectively. Then we compute the image-space gradients of the depth image $\nabla_{(u, v)}\mathcal{D}$ and finally perform the cross product of the gradients, resulting in $\mathcal{N_D}$. The depth regularized loss term on normals is defined as $L_N = 1 - \mathcal{N_D} \cdot \mathcal{N}$.
  Although there are other methods for improving normals, since our primary goal is to introduce a normal-involved shading method, we find this self-regularizer is enough for our models. Other methods are welcome to be combined with our strategy for future improvements.

  \paragraph{Capturing specular effects.}
  Our derivation on Sec.~\ref{method:oi} is based on the Lambertian assumption. To make our framework more generalizable, similar to our parameterization of the diffuse term, we aim to express the specular term, $L_s$, as a function of the normal vector $\vect{n}$. Unlike view-independent diffuse colors, specular reflectance is highly sensitive to the view direction due to the complex BRDF, $f_r(\vect{\omega_i}, \vect{\omega_o})$. Since normal vectors are implicitly involved in the BRDF to capture effects such as the Fresnel effect \cite{burley2012physically}, we cannot directly extract the normal vector as we did for 
  IDIVs in Sec.~\ref{method:oi}. Inspired by the Ref-NeRF \cite{verbin2022ref}, we model the specular component as a function of the reflection direction of the viewing direction with surface normals involved: $\vect{\omega_r} = 2(\vect{\omega_o}\cdot \vect{n})\vect{n}-\vect{\omega_o}$. By applying the Integrated Directional Encoding (IDE) \cite{verbin2022ref} on the reflection direction, we have $L_\textrm{S} = \theta(\phi_\textrm{IDE}(\vect{\omega_\textrm{r}}), \vect{n}, \vect{f}_v)$.
  The final color of the Gaussian is the summation of $L_\textrm{D}$ and $L_\textrm{S}$. More details can be found in the Appendix.

  \paragraph{Model architecture.} \label{method:model}
  Figure.~\ref{fig:method} illustrates the whole process of our method. We follow \cite{scaffoldgs} to define a set of anchor Gaussians $\{ v\}$, where each $v$ is associated with a position $\vect{x}_v$, a local feature $\vect{f}_v$, a scaling factor $\vect{s}_v$ and $k$ offsets ${\vect{O}_v^k}$ for $k$ nearby 3D Gaussians. Anchor-wise features together with global MLPs are employed to predict attributes of 3D Gaussians. We utilize our normal-involved rendering to compute surface colors $c=L_\textrm{D} + L_\textrm{S}$. Colors are finally gathered following the traditional 3DGS pipeline, together with rendered normals and depths for self-regularization. The final loss $\mathcal{L}$ consists of the original photo-metric loss $\mathcal{L}_\textrm{P}$ used in 3DGS \cite{3DGS_2023}, the volume regularization loss $\mathcal{L}_\textrm{vol}$ used in Scaffold-GS \cite{scaffoldgs} and the self-regularized depth-normal loss $\mathcal{L}_\mathcal{N}$:
  \begin{equation}\label{eq:loss}
  \mathcal{L} = \mathcal{L}_\textrm{P} + \lambda_\textrm{vol} \mathcal{L}_\textrm{vol} + \lambda_\mathcal{N} \mathcal{L}_\mathcal{N}.
  \end{equation}

%% file: Section/4_experiment.tex
\section{Experiments}\label{sec:exp}
We evaluated our Normal-GS method against several state-of-the-art 3DGS-based methods across multiple datasets, presenting both qualitative and quantitative results.

\paragraph{Baselines. }
We selected the following baseline methods for comparison:
1) \textit{3DGS} \cite{3DGS_2023} and \textit{Scaffold-GS} \cite{scaffoldgs}: These served as the vanilla and baseline methods.
2) \textit{GaussianShader (GShader)} \cite{jiang2023gaussianshader} and \textit{SpecGaussian} \cite{yang2024specgaussian}: These 3DGS-based methods modeled specular effects and used surface normals. Both methods computed the reflected direction $\vect{\omega_\textrm{r}}$ of the viewing direction w.r.t the normal. GShader used $\vect{\omega_\textrm{r}}$ to query a learned environment map for shading, while SpecGaussian used $\vect{\omega_\textrm{r}}$ to query the learned Spherical Gaussians.
Additionally, we designed a new baseline method based on Scaffold-GS. In the original Scaffold-GS, anchor features $\vect{f}_v$ were passed into a color MLP to predict 3D Gaussian colors. In our new baseline method, we also fed normals into the same color MLP, implicitly incorporating normal information. We referred to this baseline method as 3) \textit{ScaffoldGS w/ N}.

\paragraph{Datasets and evaluation metrics. }
We followed the original 3DGS \cite{3DGS_2023} methodology and used the NeRF Synthetic \cite{mildenhall2021nerf}, Mip-NeRF 360 \cite{barron2022mipnerf360}, Tank and Temple \cite{Knapitsch2017}, and Deep Blending \cite{DeepBlending2018} datasets to demonstrate the performance of our method.
Due to licensing issues, we tested on 7 out of 9 scenes in the Mip-NeRF 360 dataset. Following \cite{3DGS_2023,barron2022mipnerf360}, we selected every 8th image for testing and used the remaining images for training. We reported PSNR, SSIM, and LPIPS to measure rendering quality and the Mean Angular Error (MAE) of normals on the NeRF Synthetic \cite{mildenhall2021nerf} dataset.

\paragraph{Implementation details. }
We implemented our method in Python using the PyTorch framework. Our rasterization pipeline followed the CUDA-based rasterizer described in \cite{3DGS_2023}. We trained our models for 30k iterations, following the settings of baseline methods. Consistent with \cite{3DGS_2023, scaffoldgs}, we set $\lambda_\textrm{vol} = 0.001$. For the depth-normal loss, we used $\lambda_\mathcal{N}=0.01$. Because the depth and normals were inaccurate at the start of training, we added the depth-normal loss after training for 5k iterations. We will release our code after publication.
More details were included in the Appendix.

We tested our method and the baseline methods using their original released implementations with default hyperparameters on an NVIDIA RTX 3090 Ti GPU with 24 GB of memory. To obtain normals for comparisons, we retrained most of the baseline methods, except for 3DGS, which provided its optimized data. For methods without defined normals, such as 3DGS \cite{3DGS_2023} and Scaffold-GS \cite{scaffoldgs}, we used the shortest axis of 3D Gaussians as normals for evaluation, consistent with ours.

\subsection{Results}

\begin{figure}
    \centering
    \includegraphics[width=\linewidth]{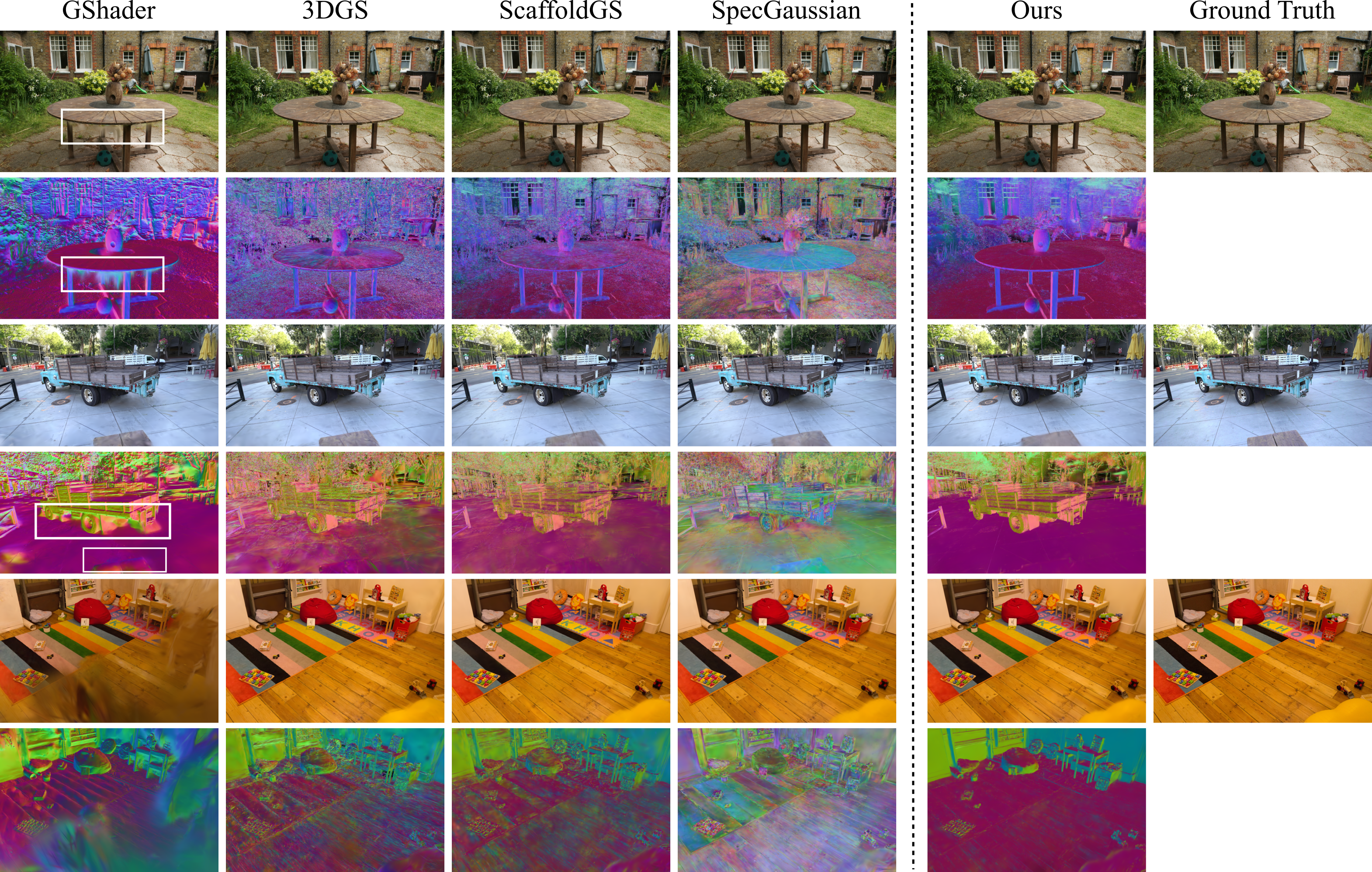}
    \caption{\textbf{Qualitative comparisons of the rendering quality and normal estimation.} Our method produced clean normals estimation and preserved good rendering quality.}
    \label{fig:main}
\end{figure}

\begin{table} \scriptsize
    \caption{\textbf{Quantitative comparisons of the rendering quality.} Our methods achieve comparable or even better results compared with the SOTA approach, SpecGaussian~\cite{yang2024specgaussian}. However, SpecGaussian performs much worse in normal estimation, while ours achieves a good balance between geometry and appearance quality. \textsuperscript{\textdagger} denotes the results quoted from their paper. GShader* fails on the Deep Blending scene. }
    \label{tab:main}
    \centering
    \begin{tabular}{llll lll lll}
        \toprule
        &  \multicolumn{3}{c}{ Mip-NeRF360} &  \multicolumn{3}{c}{ Tanks \& Temples}&  \multicolumn{3}{c}{ Deep Blending} \\
        \cmidrule(lr){2-4}  \cmidrule(lr){5-7}  \cmidrule{8-10}
        Method
        & PSNR $\uparrow$    & SSIM $\uparrow$   &  LPIPS $\downarrow$
        & PSNR $\uparrow$    & SSIM $\uparrow$   &  LPIPS $\downarrow$
        & PSNR $\uparrow$    & SSIM $\uparrow$   &  LPIPS $\downarrow$ \\
        \midrule
        3DGS\textsuperscript{\textdagger}
        & 28.691 & \cellcolor{red!40}0.870 & \cellcolor{red!40}0.182
        & 23.142 & 0.840 & 0.183
        & 29.405 & 0.903 & \cellcolor{red!40}0.243 \\
        ScaffoldGS
        & 29.267 & \cellcolor{orange!40}0.869 & \cellcolor{orange!40}0.190
        & 24.088 & 0.851 & \cellcolor{orange!40}0.175
        & 30.140 & \cellcolor{orange!40}0.905 & 0.256 \\
        ScaffoldGS w/ N
        & 29.177 & \cellcolor{red!40}0.870 & 0.192
        & 23.976 & 0.852 & 0.176
        & \cellcolor{orange!40}30.163 & \cellcolor{orange!40}0.905 & 0.263 \\
        GShader
        & 26.060 & 0.825 & 0.233   & 21.262 & 0.785 & 0.254
        & 19.159* & 0.769* & 0.453* \\
        SpecGaussian
        & \cellcolor{orange!40}29.287 & 0.864 & 0.196   & \cellcolor{red!40}24.502 & \cellcolor{red!40}0.855 & \cellcolor{orange!40}0.175
        & 30.114 &\cellcolor{orange!40}0.905 & \cellcolor{orange!40}0.252 \\
        \midrule
        Normal-GS (Ours)
        & \cellcolor{red!40}29.341 & \cellcolor{orange!40}0.869 & 0.194   & \cellcolor{orange!40}24.219 & \cellcolor{orange!40}0.854 & \cellcolor{red!40}0.174
        & \cellcolor{red!40}30.187 & \cellcolor{red!40}0.910 & \cellcolor{orange!40}0.252 \\
        \bottomrule
      \end{tabular}
      \vspace{-3mm}
  \end{table}

We first present a comprehensive comparison of rendering quality and normal estimation in Fig.\ref{fig:main}, demonstrating that our method either matches or surpasses state-of-the-art (SOTA) approaches in metrics of PSNR/SSIM/LPIPS (Tab.\ref{tab:main}). Notably, our approach significantly outperforms the baseline method (ScaffoldGS) and its variant incorporating normals (ScaffoldGS w/ N) in almost all metrics. This comparison underscores the limitations of the implicit normal utilization in ScaffoldGS w/N, which fails to accurately model appearance. In contrast, our method adopts an explicit modeling of color as a function of normals, directly derived from the rendering equation. This explicit integration enhances both the forward and backward passes during training, providing clearer cues for the network to capture nuanced color details, thereby improving rendering quality.

\begin{figure}
    \centering
    \includegraphics[width=\linewidth]{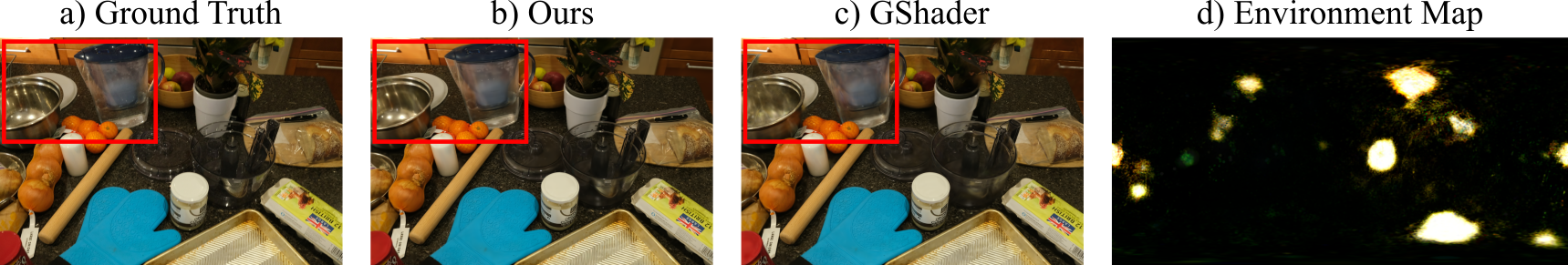}
    \caption{b) Our method successfully captures specular effects. c) GShader \cite{jiang2023gaussianshader} relying on an environment map performs poorly when d) its learned environment map fails. Scenes are taken from the Mip-NeRF 360 dataset~\cite{barron2022mipnerf360}, which have complicated lighting conditions.}
    \label{fig:environment}
    \vspace{-5mm}
  \end{figure}

Our approach also demonstrates superior rendering performance compared to methods like GShader~\cite{jiang2023gaussianshader}, which depend on global environment maps to model specular effects, as evidenced in Tab~\ref{tab:main}. The dependency of normal gradients on environment map quality, described by the equation $\frac{dL}{dN} = \frac{dL}{dE} \frac{dE}{dN}$ \cite{jiang2023gaussianshader,wu2024deferredgs}, often leads to inaccuracies in capturing detailed information, as shown in Fig.\ref{fig:environment} and further evidenced by performance issues on challenging datasets like DeepBlending\cite{DeepBlending2018} in Tab.~\ref{tab:main}. Our framework's capability to learn local incident light field attributes tied to distinct anchors in 3D space allows for more effective handling of complex lighting effects.

Although SpecGaussian~\cite{yang2024specgaussian} achieves competitive rendering quality through the use of Spherical Gaussian encoding and neural networks, it falls short in accurate normal modeling, as depicted in Fig.\ref{fig:main}. This suggests the insufficiency of relying solely on neural networks for integrating normal information into geometric modeling. Our method, with its explicit integration of normals into the rendering process, not only maintains high rendering quality but also ensures the accuracy and smoothness of normals. Our approach underscores the crucial role of direct geometric involvement in the rendering pipeline for achieving both high geometric accuracy and visual fidelity. More results are available in Appendix.

To quantitatively evaluate normal estimation, we conducted tests on the Synthetic-NeRF dataset, which includes ground-truth normal data. As shown in Tab~\ref{tab:normal}(a), our method outperforms others in the normal Mean Angular Error (MAE) metric. Visual results in Tab~\ref{tab:normal}(b) confirm our method's superior capability in capturing accurate geometry alongside high-fidelity rendering, as highlighted in the yellow box, showcasing the effectiveness of our design.

\begin{table}
	\centering
    \caption{
    \textbf{Comparison on Synthetic-NeRF dataset.} (a) The numerical results on normal consistency. Our method achieved state-of-the-art results among all methods. (b) We visualized the rendered normals for qualitative comparison. Our framework produced the best geometry.}
    \label{tab:normal}
	\setlength{\tabcolsep}{0pt}
        \begin{tabular}{cc}
    	\resizebox{0.4\textwidth}{!}{
    	    \begin{tabular}{lcc}
                \toprule
                 &Normal MAE $\downarrow$\\
                \hline
                SpecGaussian~\cite{yang2024specgaussian} & 45.98 \\
                ScaffoldGS~\cite{scaffoldgs} & 25.56 \\
                GShader~\cite{jiang2023gaussianshader} & 23.56 \\
                \hline
                Normal-GS (Ours) & \cellcolor{red!23}20.71\\
                \bottomrule
            \end{tabular}
    	}
    	&
    	\begin{minipage}{0.6\textwidth}
    		\includegraphics[width=\textwidth]{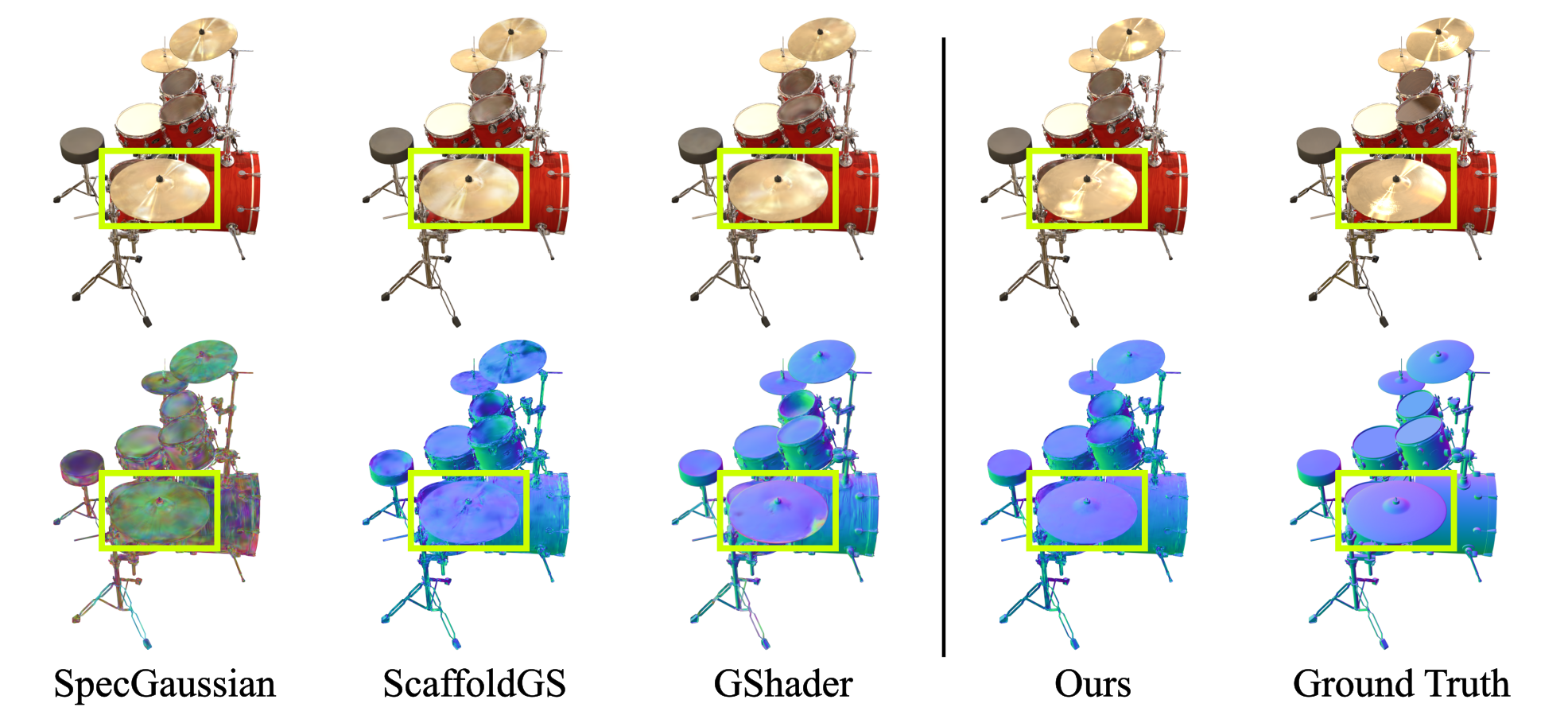}
    	\end{minipage}\\
    (a) Normal consistency comparision\small & (b) Visualization of rendered normals 	\\
    \end{tabular}
\end{table}

\subsection{Ablation studies}
  \begin{figure}[ht]
    \centering
    \includegraphics[width=\linewidth]{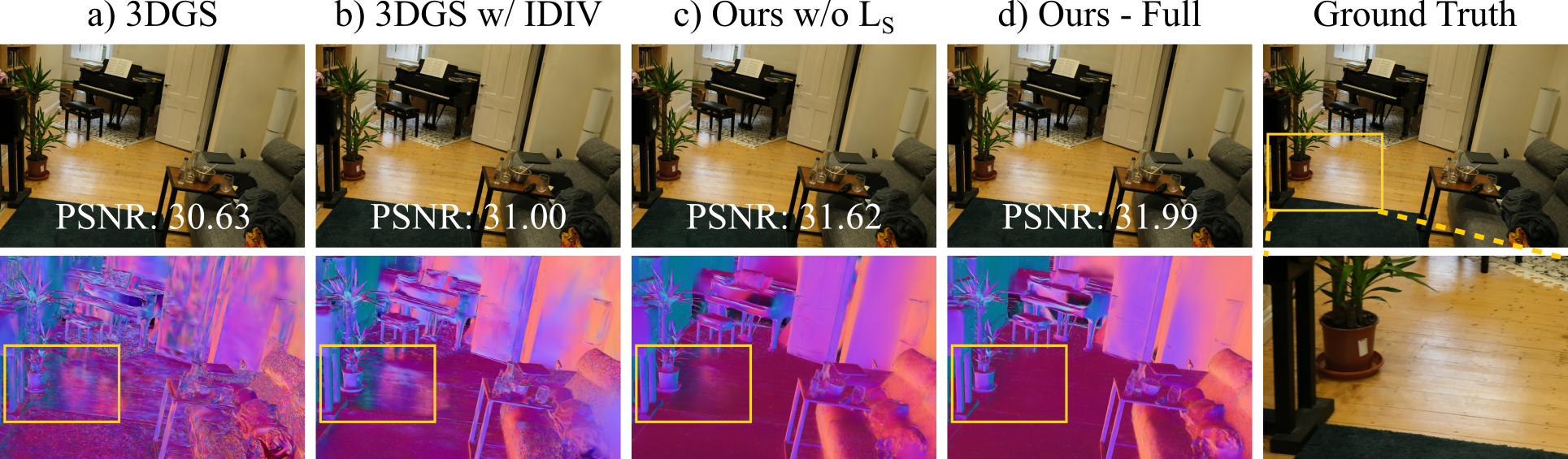}
    \caption{\textbf{Ablation study about our proposed components.} The IDIV together with our regularization strategy produces better quality in rendering quality and normal accuracy.}
    \label{fig:ablation1}
    \vspace{-3mm}
  \end{figure}
Our contributions are significant in two key areas: the integration of Integrated Directional Illumination Vectors (IDIV) and the implementation of an anchor-based regularization design. We initially validate the impact of IDIV through an ablation study conducted within a conventional 3DGS framework, detailed in Sec.~\ref{method:oi}. Introducing IDIV necessitates additional parameters, requiring the adoption of regularization techniques such as Laplacian and Total-Variation loss, referenced in~\cite{xu2017shading}. The outcomes, illustrated in Fig.~\ref{fig:ablation1} (a) and (b), demonstrate substantial enhancements in rendering quality and the smoothing of normals.

However, we found that the model's performance is highly sensitive to the tuning of loss weights. To address this, we implemented an anchor-based IDIV regularization strategy in Sec.~\ref{method:local} with results illustrated in Fig.~\ref{fig:ablation1} (c). This approach employs neural networks to model locally shared IDIVs, proving to be both robust and efficient in regularizing the solution space. Notably, the simplicity yet effectiveness of our final loss term, defined in Eq.~\ref{eq:loss}, underscores the potential of our design to enhance both geometric and rendering quality. When considering the specular component, our full (Fig.~\ref{fig:ablation1} (d)) model achieves superior results. 

%% file: Section/5_conclusion.tex
\section{Discussions}\label{sec:discussion}
\textbf{Conclusion.} We introduce a novel shading technique within the 3D Gaussian Splatting (3DGS) framework to improve view synthesis and normal estimation. By integrating Directional Illumination Vectors (IDIV) as advanced attributes for color representation and utilizing an anchor-based design with a neural MLP for IDIV prediction, our approach achieves a superior balance between rendering quality and geometric accuracy. This enhancement not only elevates image fidelity but also ensures precise normal estimations, pushing the boundaries of 3DGS technology in real-world applications.

\textbf{Limitation and Future Work.} We acknowledge existing limitations within our framework. Notably, the self-regularized normal loss proves suboptimal for certain outdoor scenes at a distance (as observed in the Appendix~\ref{fig:limitation}), negatively impacting rendering quality in these regions. This issue could potentially be addressed with more sophisticated normal guidance techniques. In future research, we plan to delve into a more granular decomposition of the Integrated Directional Illumination Vectors (IDIV) and specular components. This exploration aims to enhance capabilities in texture editing and relighting, thereby broadening the practical applications of our framework.

\textbf{Boarder Impact.} As our approaches require per-scene optimization to obtain the 3D model for each scene, the computational resources for model training could be a concern for global climate change.